# A System for Induction of Oblique Decision Trees

**Sreerama K. Murthy** MURTHY@CS.JHU.EDU
**Simon Kasif** KASIF@CS.JHU.EDU
**Steven Salzberg** SALZBERG@CS.JHU.EDU
*Department of Computer Science*
*Johns Hopkins University, Baltimore, MD 21218 USA*

## Abstract

This article describes a new system for induction of oblique decision trees. This system, OC1, combines deterministic hill-climbing with two forms of randomization to find a good oblique split (in the form of a hyperplane) at each node of a decision tree. Oblique decision tree methods are tuned especially for domains in which the attributes are numeric, although they can be adapted to symbolic or mixed symbolic/numeric attributes. We present extensive empirical studies, using both real and artificial data, that analyze OC1's ability to construct oblique trees that are smaller and more accurate than their axis-parallel counterparts. We also examine the benefits of randomization for the construction of oblique decision trees.

## 1. Introduction

Current data collection technology provides a unique challenge and opportunity for automated machine learning techniques. The advent of major scientific projects such as the Human Genome Project, the Hubble Space Telescope, and the human brain mapping initiative are generating enormous amounts of data on a daily basis. These streams of data require automated methods to analyze, filter, and classify them before presenting them in digested form to a domain scientist. Decision trees are a particularly useful tool in this context because they perform classification by a sequence of simple, easy-to-understand tests whose semantics is intuitively clear to domain experts. Decision trees have been used for classification and other tasks since the 1960s (Moret, 1982; Safavin & Landgrebe, 1991). In the 1980's, Breiman et al.'s book on classification and regression trees (CART) and Quinlan's work on ID3 (Quinlan, 1983, 1986) provided the foundations for what has become a large body of research on one of the central techniques of experimental machine learning.

Many variants of decision tree (DT) algorithms have been introduced in the last decade. Much of this work has concentrated on decision trees in which each node checks the value of a single attribute (Breiman, Friedman, Olshen, & Stone, 1984; Quinlan, 1986, 1993a). Quinlan initially proposed decision trees for classification in domains with symbolic-valued attributes (1986), and later extended them to numeric domains (1987). When the attributes are numeric, the tests have the form $x_i > k$, where $x_i$ is one of the attributes of an example and $k$ is a constant. This class of decision trees may be called *axis-parallel*, because the tests at each node are equivalent to axis-parallel hyperplanes in the attribute space. An example of such a decision tree is given in Figure 1, which shows both a tree and the partitioning it creates in a 2-D attribute space.

Figure 1: The left side of the figure shows a simple axis-parallel tree that uses two attributes. The right side shows the partitioning that this tree creates in the attribute space.

Researchers have also studied decision trees in which the test at a node uses boolean combinations of attributes (Pagallo, 1990; Pagallo & Haussler, 1990; Sahami, 1993) and linear combinations of attributes (see Section 2). Different methods for measuring the goodness of decision tree nodes, as well as techniques for pruning a tree to reduce *overfitting* and increase accuracy have also been explored, and will be discussed in later sections.

In this paper, we examine decision trees that test a linear combination of the attributes at each internal node. More precisely, let an example take the form $X = x_1, x_2, \ldots, x_d, C_j$ where $C_j$ is a class label and the $x_i$'s are real-valued attributes.[1] The test at each node will then have the form:

$$\sum_{i=1}^{d} a_i x_i + a_{d+1} > 0 \qquad (1)$$

where $a_1, \ldots, a_{d+1}$ are real-valued coefficients. Because these tests are equivalent to hyperplanes at an oblique orientation to the axes, we call this class of decision trees *oblique* decision trees. (Trees of this form have also been called "multivariate" (Brodley & Utgoff, 1994). We prefer the term "oblique" because "multivariate" includes non-linear combinations of the variables, i.e., curved surfaces. Our trees contain only linear tests.) It is clear that these are simply a more general form of axis-parallel trees, since by setting $a_i = 0$ for all coefficients but one, the test in Eq. 1 becomes the familiar univariate test. Note that oblique decision trees produce polygonal (polyhedral) partitionings of the attribute space, while axis-parallel trees produce partitionings in the form of hyper-rectangles that are parallel to the feature axes.

It should be intuitively clear that when the underlying concept is defined by a polygonal space partitioning, it is preferable to use oblique decision trees for classification. For example, there exist many domains in which one or two oblique hyperplanes will be the best model to use for classification. In such domains, axis-parallel methods will have to ap-

1. The constraint that $x_1, \ldots, x_d$ are real-valued does not necessarily restrict oblique decision trees to numeric domains. Several researchers have studied the problem of converting symbolic (unordered) domains to numeric (ordered) domains and vice versa; e.g., (Breiman et al., 1984; Hampson & Volper, 1986; Utgoff & Brodley, 1990; Van de Merckt, 1992, 1993). To keep the discussion simple, however, we will assume that all attributes have numeric values.

Figure 2: The left side shows a simple 2-D domain in which two oblique hyperplanes define the classes. The right side shows an approximation of the sort that an axis-parallel decision tree would have to create to model this domain.

proximate the correct model with a staircase-like structure, while an oblique tree-building method could capture it with a tree that was both smaller and more accurate.[2] Figure 2 gives an illustration.

Breiman et al. first suggested a method for inducing oblique decision trees in 1984. However, there has been very little further research on such trees until relatively recently (Utgoff & Brodley, 1990; Heath, Kasif, & Salzberg, 1993b; Murthy, Kasif, Salzberg, & Beigel, 1993; Brodley & Utgoff, 1994). A comparison of existing approaches is given in more detail in Section 2. The purpose of this study is to review the strengths and weaknesses of existing methods, to design a system that combines some of the strengths and overcomes the weaknesses, and to evaluate that system empirically and analytically. The main contributions and conclusions of our study are as follows:

- We have developed a new, randomized algorithm for inducing oblique decision trees from examples. This algorithm extends the original 1984 work of Breiman et al. Randomization helps significantly in learning many concepts.

- Our algorithm is fully implemented as an oblique decision tree induction system and is available over the Internet. The code can be retrieved from Online Appendix 1 of this paper (or by anonymous ftp from ftp://ftp.cs.jhu.edu/pub/oc1/oc1.tar.Z).

- The randomized hill-climbing algorithm used in OC1 is more efficient than other existing randomized oblique decision tree methods (described below). In fact, the current implementation of OC1 guarantees a worst-case running time that is only $O(\log n)$ times greater than the worst-case time for inducing axis-parallel trees (i.e., $O(dn^2 \log n)$ vs. $O(dn^2)$).

- The ability to generate oblique trees often produces very small trees compared to axis-parallel methods. When the underlying problem requires an oblique split, oblique

2. Note that though a given oblique tree may have fewer leaf nodes than an axis-parallel tree—which is what we mean by "smaller"—the oblique tree may in some cases be larger in terms of information content, because of the increased complexity of the tests at each node.

trees are also more accurate than axis-parallel trees. Allowing a tree-building system to use both oblique and axis-parallel splits broadens the range of domains for which the system should be useful.

The remaining sections of the paper follow this outline: the remainder of this section briefly outlines the general paradigm of decision tree induction, and discusses the complexity issues involved in inducing oblique decision trees. Section 2 briefly reviews some existing techniques for oblique DT induction, outlines some limitations of each approach, and introduces the OC1 system. Section 3 describes the OC1 system in detail. Section 4 describes experiments that (1) compare the performance of OC1 to that of several other axis-parallel and oblique decision tree induction methods on a range of real-world datasets and (2) demonstrate empirically that OC1 significantly benefits from its randomization methods. In Section 5, we conclude with some discussion of open problems and directions for further research.

### 1.1 Top-Down Induction of Decision Trees

Algorithms for inducing decision trees follow an approach described by Quinlan as top-down induction of decision trees (1986). This can also be called a greedy divide-and-conquer method. The basic outline is as follows:

1. Begin with a set of examples called the training set, $T$. If all examples in $T$ belong to one class, then halt.
2. Consider all tests that divide $T$ into two or more subsets. Score each test according to how well it splits up the examples.
3. Choose ("greedily") the test that scores the highest.
4. Divide the examples into subsets and run this procedure recursively on each subset.

Quinlan's original model only considered attributes with symbolic values; in that model, a test at a node splits an attribute into all of its values. Thus a test on an attribute with three values will have at most three child nodes, one corresponding to each value. The algorithm considers *all* possible tests and chooses the one that optimizes a pre-defined goodness measure. (One could also split symbolic values into two or more subsets of values, which gives many more choices for how to split the examples.) As we explain next, oblique decision tree methods cannot consider all tests due to complexity considerations.

### 1.2 Complexity of Induction of Oblique Decision Trees

One reason for the relatively few papers on the problem of inducing oblique decision trees is the increased computational complexity of the problem when compared to the axis-parallel case. There are two important issues that must be addressed. In the context of top-down decision tree algorithms, we must address the complexity of finding optimal separating hyperplanes (decision surfaces) for a given node of a decision tree. An optimal hyperplane will minimize the impurity measure used; e.g., impurity might be measured by the total number of examples mis-classified. The second issue is the lower bound on the complexity of finding optimal (e.g., smallest size) trees.

Figure 3: For $n$ points in $d$ dimensions ($n \geq d$), there are $n \cdot d$ distinct axis-parallel splits, while there are $2^d \cdot \binom{n}{d}$ distinct $d$-dimensional oblique splits. This shows all distinct oblique and axis-parallel splits for two specific points in 2-D.

Let us first consider the issue of the complexity of selecting an optimal oblique hyperplane for a single node of a tree. In a domain with $n$ training instances, each described using $d$ real-valued attributes, there are at most $2^d \cdot \binom{n}{d}$ distinct $d$-dimensional oblique splits; i.e., hyperplanes[3] that divide the training instances uniquely into two nonoverlapping subsets. This upper bound derives from the observation that every subset of size $d$ from the $n$ points can define a $d$-dimensional hyperplane, and each such hyperplane can be rotated slightly in $2^d$ directions to divide the set of $d$ points in all possible ways. Figure 3 illustrates these upper limits for two points in two dimensions. For axis-parallel splits, there are only $n \cdot d$ distinct possibilities, and axis-parallel methods such as C4.5 (Quinlan, 1993a) and CART (Breiman et al., 1984) can exhaustively search for the best split at each node. The problem of searching for the best oblique split is therefore much more difficult than that of searching for the best axis-parallel split. In fact, the problem is NP-hard.

More precisely, Heath (1992) proved that the following problem is NP-hard: given a set of labelled examples, find the hyperplane that minimizes the number of misclassified examples both above and below the hyperplane. This result implies that any method for finding the optimal oblique split is likely to have exponential cost (assuming $P \neq NP$). Intuitively, the problem is that it is impractical to enumerate all $2^d \cdot \binom{n}{d}$ distinct hyperplanes and choose the best, as is done in axis-parallel decision trees. However, any non-exhaustive deterministic algorithm for searching through all these hyperplanes is prone to getting stuck in local minima.

3. Throughout the paper, we use the terms "split" and "hyperplane" interchangeably to refer to the test at a node of a decision tree. The first usage is standard (Moret, 1982), and refers to the fact that the test splits the data into two partitions. The second usage refers to the geometric form of the test.

On the other hand, it is possible to define impurity measures for which the problem of finding optimal hyperplanes can be solved in polynomial time. For example, if one minimizes the sum of distances of mis-classified examples, then the optimal solution can be found using linear programming methods (if distance is measured along one dimension only). However, classifiers are usually judged by how many points they classify correctly, regardless of how close to the decision boundary a point may lie. Thus most of the standard measures for computing impurity base their calculation on the discrete number of examples of each category on either side of the hyperplane. Section 3.3 discusses several commonly used impurity measures.

Now let us address the second issue, that of the complexity of building a small tree. It is easy to show that the problem of inducing the smallest axis-parallel decision tree is NP-hard. This observation follows directly from the work of Hyafil and Rivest (1976). Note that one can generate the smallest axis-parallel tree that is consistent with the training set in polynomial time *if* the number of attributes is a constant. This can be done by using dynamic programming or branch and bound techniques (see Moret (1982) for several pointers). But when the tree uses oblique splits, it is not clear, even for a fixed number of attributes, how to generate an optimal (e.g., smallest) decision tree in polynomial time. This suggests that the complexity of constructing good oblique trees is greater than that for axis-parallel trees.

It is also easy to see that the problem of constructing an optimal (e.g., smallest) oblique decision tree is NP-hard. This conclusion follows from the work of Blum and Rivest (1988). Their result implies that in $d$ dimensions (i.e., with $d$ attributes) the problem of producing a 3-node oblique decision tree that is consistent with the training set is NP-complete. More specifically, they show that the following decision problem is NP-complete: given a training set $T$ with $n$ examples and $d$ Boolean attributes, does there exist a 3-node neural network consistent with $T$? From this it is easy to show that the following question is NP-complete: given a training set $T$, does there exist a 3-leaf-node oblique decision tree consistent with $T$?

As a result of these complexity considerations, we took the pragmatic approach of trying to generate small trees, but not looking for the smallest tree. The greedy approach used by OC1 and virtually all other decision tree algorithms implicitly tries to generate small trees. In addition, it is easy to construct example problems for which the optimal split at a node will not lead to the best tree; thus our philosophy as embodied in OC1 is to find locally good splits, but not to spend excessive computational effort on improving the quality of these splits.

## 2. Previous Work on Oblique Decision Tree Induction

Before describing the OC1 algorithm, we will briefly discuss some existing oblique DT induction methods, including CART with linear combinations, Linear Machine Decision Trees, and Simulated Annealing of Decision Trees. There are also methods that induce tree-like classifiers with linear discriminants at each node, most notably methods using linear programming (Mangasarian, Setiono, & Wolberg, 1990; Bennett & Mangasarian, 1992, 1994a, 1994b). Though these methods can find the optimal linear discriminants for specific goodness measures, the size of the linear program grows very fast with the number

**To induce a split at node $T$ of the decision tree:**
  **Normalize values for all $d$ attributes.**
  $L = 0$
  **While (TRUE)**
    $L = L + 1$
    **Let the current split $s_L$ be $v \leq c$, where $v = \sum_{i=1}^{d} a_i x_i$.**
    **For $i = 1, \ldots, d$**
      **For $\gamma$ = -0.25,0,0.25**
        **Search for the $\delta$ that maximizes the goodness of the split $v - \delta(a_i + \gamma) \leq c$.**
      **Let $\delta^*$,$\gamma^*$ be the settings that result in highest goodness in these 3 searches.**
      $a_i = a_i - \delta^*$, $c = c - \delta^* \gamma^*$.
    **Perturb $c$ to maximize the goodness of $s_L$, keeping $a_1, \ldots, a_d$ constant.**
    **If |goodness($s_L$) - goodness($s_{L-1}$)| $\leq \epsilon$ exit while loop.**
  **Eliminate irrelevant attributes in $\{a_1, \ldots, a_d\}$ using backward elimination.**
  **Convert $s_L$ to a split on the un-normalized attributes.**
  **Return the better of $s_L$ and the best axis-parallel split as the split for $T$.**

Figure 4: The procedure used by CART with linear combinations (CART-LC) at each node of a decision tree.

of instances and the number of attributes. There is also some less closely related work on algorithms to train artificial neural networks to build decision tree-like classifiers (Brent, 1991; Cios & Liu, 1992; Herman & Yeung, 1992).

The first oblique decision tree algorithm to be proposed was CART with linear combinations (Breiman et al., 1984, chapter 5). This algorithm, referred to henceforth as CART-LC, is an important basis for OC1. Figure 4 summarizes (using Breiman et al.'s notation) what the CART-LC algorithm does at each node in the decision tree. The core idea of the CART-LC algorithm is how it finds the value of $\delta$ that maximizes the goodness of a split. This idea is also used in OC1, and is explained in detail in Section 3.1.

After describing CART-LC, Breiman et al. point out that there is still much room for further development of the algorithm. OC1 represents an extension of CART-LC that includes some significant additions. It addresses the following limitations of CART-LC:

- CART-LC is fully deterministic. There is no built-in mechanism for escaping local minima, although such minima may be very common for some domains. Figure 5 shows a simple example for which CART-LC gets stuck.
- CART-LC produces only a single tree for a given data set.
- CART-LC sometimes makes adjustments that increase the impurity of a split. This feature was probably included to allow it to escape some local minima.
- There is no upper bound on the time spent at any node in the decision tree. It halts when no perturbation changes the impurity more than $\epsilon$, but because impurity may increase and decrease, the algorithm can spend arbitrarily long time at a node.

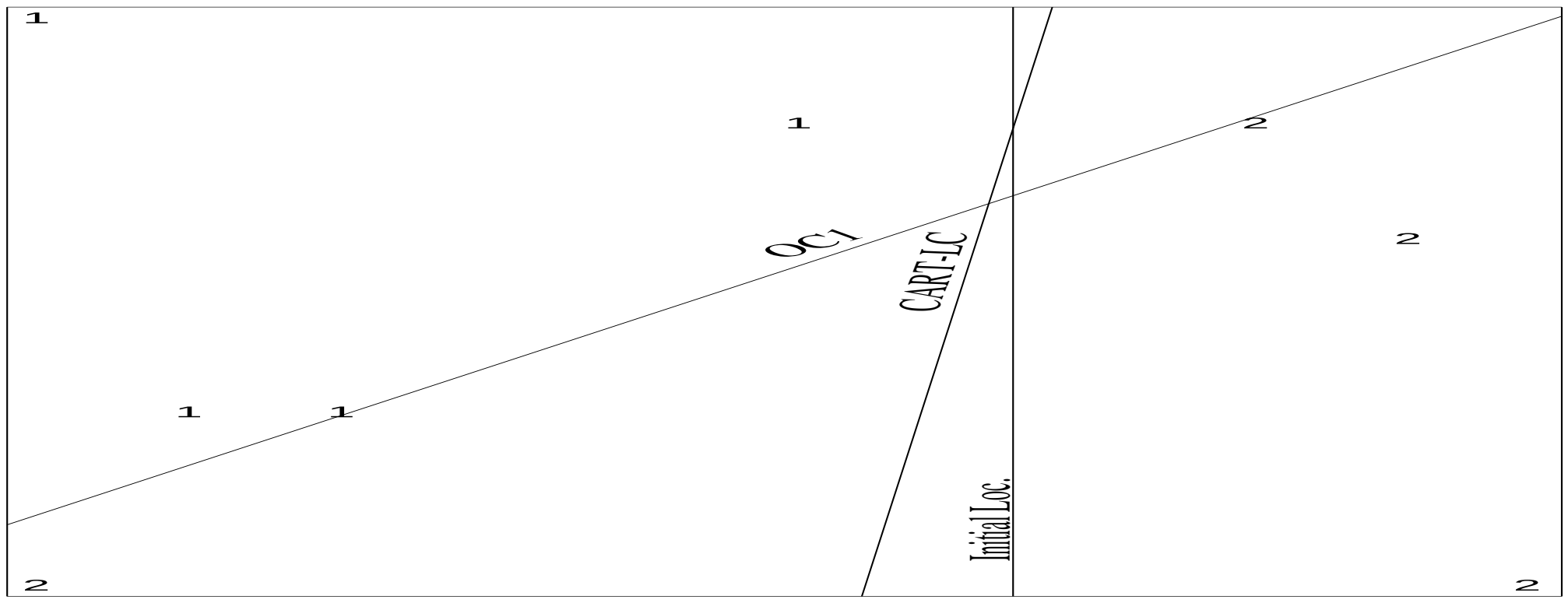

Figure 5: The deterministic perturbation algorithm of CART-LC fails to find the correct split for this data, even when it starts from the location of the best axis-parallel split. OC1 finds the correct split using one random jump.

Another oblique decision tree algorithm, one that uses a very different approach from CART-LC, is the Linear Machine Decision Trees (LMDT) system (Utgoff & Brodley, 1991; Brodley & Utgoff, 1992), which is a successor to the Perceptron Tree method (Utgoff, 1989; Utgoff & Brodley, 1990). Each internal node in an LMDT tree is a Linear Machine (Nilsson, 1990). The training algorithm presents examples repeatedly at each node until the linear machine converges. Because convergence cannot be guaranteed, LMDT uses heuristics to determine when the node has stabilized. To make the training stable even when the set of training instances is not linearly separable, a "thermal training" method (Frean, 1990) is used, similar to simulated annealing.

A third system that creates oblique trees is Simulated Annealing of Decision Trees (SADT) (Heath et al., 1993b) which, like OC1, uses randomization. SADT uses simulated annealing (Kirkpatrick, Gelatt, & Vecci, 1983) to find good values for the coefficients of the hyperplane at each node of a tree. SADT first places a hyperplane in a canonical location, and then iteratively perturbs all the coefficients by small random amounts. Initially, when the temperature parameter is high, SADT accepts almost any perturbation of the hyperplane, regardless of how it changes the goodness score. However, as the system "cools down," only changes that improve the goodness of the split are likely to be accepted. Though SADT's use of randomization allows it to effectively avoid some local minima, it compromises on efficiency. It runs much slower than either CART-LC, LMDT or OC1, sometimes considering tens of thousands of hyperplanes at a single node before it finishes annealing.

Our experiments in Section 4.3 include some results showing how all of these methods perform on three artificial domains.

We next describe a way to combine some of the strengths of the methods just mentioned, while avoiding some of the problems. Our algorithm, OC1, uses deterministic hill climbing most of the time, ensuring computational efficiency. In addition, it uses two kinds of randomization to avoid local minima. By limiting the number of random choices, the algorithm is guaranteed to spend only polynomial time at each node in the tree. In addition, randomization itself has produced several benefits: for example, it means that the algorithm

**To find a split of a set of examples $T$:**
**Find the best axis-parallel split of $T$. Let $I$ be the impurity of this split.**
**Repeat $R$ times:**
**Choose a random hyperplane $H$.**
**(For the first iteration, initialize $H$ to be the best axis-parallel split.)**
**Step 1: Until the impurity measure does not improve, do:**
**Perturb each of the coefficients of $H$ in sequence.**
**Step 2: Repeat at most $J$ times:**
**Choose a random direction and attempt to perturb $H$ in that direction.**
**If this reduces the impurity of $H$, go to Step 1.**
**Let $I_1$ = the impurity of $H$. If $I_1 < I$, then set $I = I_1$.**
**Output the split corresponding to $I$.**

Figure 6: Overview of the OC1 algorithm for a single node of a decision tree.

can produce many different trees for the same data set. This offers the possibility of a new family of classifiers: $k$-decision-tree algorithms, in which an example is classified by the majority vote of $k$ trees. Heath et al. (1993a) have shown that $k$-decision tree methods (which they call $k$-DT) will consistently outperform single tree methods if classification accuracy is the main criterion. Finally, our experiments indicate that OC1 efficiently finds small, accurate decision trees for many different types of classification problems.

## 3. Oblique Classifier 1 (OC1)

In this section we discuss details of the oblique decision tree induction system OC1. As part of this description, we include:

- the method for finding coefficients of a hyperplane at each tree node,
- methods for computing the impurity or goodness of a hyperplane,
- a tree pruning strategy, and
- methods for coping with missing and irrelevant attributes.

Section 3.1 focuses on the most complicated of these algorithmic details; i.e. the question of how to find a hyperplane that splits a given set of instances into two reasonably "pure" non-overlapping subsets. This randomized perturbation algorithm is the main novel contribution of OC1. Figure 6 summarizes the basic OC1 algorithm, used at each node of a decision tree. This figure will be explained further in the following sections.

### 3.1 Perturbation algorithm

OC1 imposes no restrictions on the orientation of the hyperplanes. However, in order to be at least as powerful as standard DT methods, it first finds the best axis-parallel (univariate) split at a node before looking for an oblique split. OC1 uses an oblique split only when it improves over the best axis-parallel split.[4]

4. As pointed out in (Breiman et al., 1984, Chapter 5), it does not make sense to use an oblique split when the number of examples at a node $n$ is less than or almost equal to the number of features $d$, because the The search strategy for the space of possible hyperplanes is defined by the procedure that perturbs the current hyperplane $H$ to a new location. Because there are an exponential number of distinct ways to partition the examples with a hyperplane, any procedure that simply enumerates all of them will be unreasonably costly. The two main alternatives considered in the past have been simulated annealing, used in the SADT system (Heath et al., 1993b), and deterministic heuristic search, as in CART-LC (Breiman et al., 1984). OC1 combines these two ideas, using heuristic search until it finds a local minimum, and then using a non-deterministic search step to get out of the local minimum. (The non-deterministic step in OC1 is *not* simulated annealing, however.)

We will start by explaining how we perturb a hyperplane to split the training set $T$ at a node of the decision tree. Let $n$ be the number of examples in $T$, $d$ be the number of attributes (or dimensions) for each example, and $k$ be the number of categories. Then we can write $T_j = (x_{j1}, x_{j2}, \ldots, x_{jd}, C_j)$ for the $j$th example from the training set $T$, where $x_{ji}$ is the value of attribute $i$ and $C_j$ is the category label. As defined in Eq. 1, the equation of the current hyperplane $H$ at a node of the decision tree is written as $\sum_{i=1}^{d}(a_i x_i) + a_{d+1} = 0$. If we substitute a point (an example) $T_j$ into the equation for $H$, we get $\sum_{i=1}^{d}(a_i x_{ji}) + a_{d+1} = V_j$, where the sign of $V_j$ tells us whether the point $T_j$ is above or below the hyperplane $H$; i.e., if $V_j > 0$, then $T_j$ is above $H$. If $H$ splits the training set $T$ perfectly, then all points belonging to the same category will have the same sign for $V_j$. i.e., $\text{sign}(V_i) = \text{sign}(V_j)$ iff $\text{category}(T_i) = \text{category}(T_j)$.

OC1 adjusts the coefficients of $H$ individually, finding a locally optimal value for one coefficient at a time. This key idea was introduced by Breiman et al. It works as follows. Treat the coefficient $a_m$ as a variable, and treat all other coefficients as constants. Then $V_j$ can be viewed as a function of $a_m$. In particular, the condition that $T_j$ is above $H$ is equivalent to

$$V_j > 0$$

$$a_m > \frac{a_m x_{jm} - V_j}{x_{jm}} \stackrel{\text{def}}{=} U_j \qquad (2)$$

assuming that $x_{jm} > 0$, which we ensure by normalization. Using this definition of $U_j$, the point $T_j$ is above $H$ if $a_m > U_j$, and below otherwise. By plugging all the points from $T$ into this equation, we will obtain $n$ constraints on the value of $a_m$.

The problem then is to find a value for $a_m$ that satisfies as many of these constraints as possible. (If all the constraints are satisfied, then we have a perfect split.) This problem is easy to solve optimally: simply sort all the values $U_j$, and consider setting $a_m$ to the midpoint between each pair of different values. This is illustrated in Figure 7. In the figure, the categories are indicated by font size; the larger $U_i$'s belong to one category, and the smaller to another. For each distinct placement of the coefficient $a_m$, OC1 computes the impurity of the resulting split; e.g., for the location between $U_6$ and $U_7$ illustrated here, two examples on the left and one example on the right would be misclassified (see Section 3.3.1 for different ways of computing impurity). As the figure illustrates, the problem is simply to find the best one-dimensional split of the $U$s, which requires considering just $n-1$ values for $a_m$. The value $a'_m$ obtained by solving this one-dimensional problem is then considered data *underfits* the concept. By default, OC1 uses only axis-parallel splits at tree nodes at which $n < 2d$. The user can vary this threshold.

Figure 7: Finding the optimal value for a single coefficient $a_m$. Large U's correspond to examples in one category and small u's to another.

**Perturb(H,m)**
**For** $j = 1, \ldots, n$
**Compute** $U_j$ **(Eq. 2)**
**Sort** $U_1, \ldots, U_n$ **in non-decreasing order.**
$a'_m$ **= best univariate split of the sorted** $U_j$**s.**
$H_1$ **= result of substituting** $a'_m$ **for** $a_m$ **in** $H$**.**
**If** $(impurity(H_1) < impurity(H))$
$\{ a_m = a'_m \; ; \; P_{move} = P_{stag} \}$
**Else if** $(impurity(H) = impurity(H_1))$
$\{ a_m = a'_m$ **with probability** $P_{move}$
$P_{move} = P_{move} - 0.1 * P_{stag} \}$

Figure 8: Perturbation algorithm for a single coefficient $a_m$.

as a replacement for $a_m$. Let $H_1$ be the hyperplane obtained by "perturbing" $a_m$ to $a'_m$. If $H$ has better (lower) impurity than $H_1$, then $H_1$ is discarded. If $H_1$ has lower impurity, $H_1$ becomes the new location of the hyperplane. If $H$ and $H_1$ have identical impurities, then $H_1$ replaces $H$ with probability $P_{stag}$.[5] Figure 8 contains pseudocode for our perturbation procedure.

Now that we have a method for locally improving a coefficient of a hyperplane, we need to decide which of the $d + 1$ coefficients to pick for perturbation. We experimented with three different methods for choosing which coefficient to adjust, namely, sequential, best first and random.

**Seq**: Repeat until none of the coefficient values is modified in the **For** loop:
For $i = 1$ to $d$, Perturb$(H, i)$

**Best**: Repeat until coefficient $m$ remains unmodified:
$m$ = coefficient which when perturbed, results in the maximum improvement of the impurity measure.
Perturb$(H, m)$

**R-50**: Repeat a fixed number of times (50 in our experiments):
$m$ = random integer between 1 and $d + 1$
Perturb$(H, m)$

5. The parameter $P_{stag}$, denoting "stagnation probability", is the probability that a hyperplane is perturbed to a location that does not change the impurity measure. To prevent the impurity from remaining stagnant for a long time, $P_{stag}$ decreases by a constant amount each time OC1 makes a "stagnant" perturbation; thus only a constant number of such perturbations will occur at each node. This constant can be set by the user. $P_{stag}$ is reset to 1 every time the global impurity measure is improved.

Our previous experiments (Murthy et al., 1993) indicated that the order of perturbation of the coefficients does not affect the classification accuracy as much as other parameters, especially the randomization parameters (see below). Since none of these orders was uniformly better than any other, we used sequential (Seq) perturbation for all the experiments reported in Section 4.

### 3.2 Randomization

The perturbation algorithm halts when the split reaches a local minimum of the impurity measure. For OC1's search space, a local minimum occurs when no perturbation of any single coefficient of the current hyperplane will decrease the impurity measure. (Of course, a local minimum may also be a global minimum.) We have implemented two ways of attempting to escape local minima: perturbing the hyperplane with a random vector, and re-starting the perturbation algorithm with a different random initial hyperplane.

The technique of perturbing the hyperplane with a random vector works as follows. When the system reaches a local minimum, it chooses a random vector to add to the coefficients of the current hyperplane. It then computes the optimal amount by which the hyperplane should be perturbed along this random direction. To be more precise, when a hyperplane $H = \sum_{i=1}^{d} a_i x_i + a_{d+1}$ cannot be improved by deterministic perturbation, OC1 repeats the following loop $J$ times (where $J$ is a user-specified parameter, set to 5 by default).

- Choose a random vector $R = (r_1, r_2, \ldots, r_{d+1})$.
- Let $\alpha$ be the amount by which we want to perturb $H$ in the direction $R$. In other words, let $H_1 = \sum_{i=1}^{d} (a_i + \alpha r_i) x_i + (a_{d+1} + \alpha r_{d+1})$.
- Find the optimal value for $\alpha$.
- If the hyperplane $H_1$ thus obtained decreases the overall impurity, replace $H$ with $H_1$, exit this loop and begin the deterministic perturbation algorithm for the individual coefficients.

Note that we can treat $\alpha$ as the only variable in the equation for $H_1$. Therefore each of the $n$ examples in $T$, if plugged into the equation for $H_1$, imposes a constraint on the value of $\alpha$. OC1 therefore can use its coefficient perturbation method (see Section 3.1) to compute the best value of $\alpha$. If $J$ random jumps fail to improve the impurity, OC1 halts and uses $H$ as the split for the current tree node.

An intuitive way of understanding this random jump is to look at the dual space in which the algorithm is actually searching. Note that the equation $H = \sum_{i=1}^{d} a_i x_i + a_{d+1}$ defines a space in which the axes are the coefficients $a_i$ rather than the attributes $x_i$. Every point in this space defines a distinct hyperplane in the original formulation. The deterministic algorithm used in OC1 picks a hyperplane and then adjusts coefficients one at a time. Thus in the dual space, OC1 chooses a point and perturbs it by moving it parallel to the axes. The random vector $R$ represents a random *direction* in this space. By finding the best value for $\alpha$, OC1 finds the best distance to adjust the hyperplane in the direction of $R$.

Note that this additional perturbation in a random direction does not significantly increase the time complexity of the algorithm (see Appendix A). We found in our experiments that even a single random jump, when used at a local minimum, proves to be very helpful. Classification accuracy improved for every one of our data sets when such perturbations were made. See Section 4.3 for some examples.

The second technique for avoiding local minima is a variation on the idea of performing multiple local searches. The technique of multiple local searches is a natural extension to local search, and has been widely mentioned in the optimization literature (see Roth (1970) for an early example). Because most of the steps of our perturbation algorithm are deterministic, the initial hyperplane largely determines which local minimum will be encountered first. Perturbing a single initial hyperplane is thus unlikely to lead to the best split of a given data set. In cases where the random perturbation method fails to escape from local minima, it may be helpful to simply start afresh with a new initial hyperplane. We use the word *restart* to denote one run of the perturbation algorithms, at one node of the decision tree, using one random initial hyperplane.[6] That is, a restart cycles through and perturbs the coefficients one at a time and then tries to perturb the hyperplane in a random direction when the algorithm reaches a local minimum. If this last perturbation reduces the impurity, the algorithm goes back to perturbing the coefficients one at a time. The restart ends when neither the deterministic local search nor the random jump can find a better split. One of the optional parameters to OC1 specifies how many restarts to use. If more than one restart is used, then the best hyperplane found thus far is always saved. In all our experiments, the classification accuracies increased with more than one restart. Accuracy tended to increase up to a point and then level off (after about 20–50 restarts, depending on the domain). Overall, the use of multiple initial hyperplanes substantially improved the quality of the decision trees found (see Section 4.3 for some examples).

By carefully combining hill-climbing and randomization, OC1 ensures a worst case time of $O(dn^2 \log n)$ for inducing a decision tree. See Appendix A for a derivation of this upper bound.

**Best Axis-Parallel Split.** It is clear that axis-parallel splits are more suitable for some data distributions than oblique splits. To take into account such distributions, OC1 computes the best axis-parallel split *and* an oblique split at each node, and then picks the better of the two.[7] Calculating the best axis-parallel split takes an additional $O(dn \log n)$ time, and so does not increase the asymptotic time complexity of OC1. As a simple variant of the OC1 system, the user can opt to "switch off" the oblique perturbations, thus building an axis-parallel tree on the training data. Section 4.2 empirically demonstrates that this axis-parallel variant of OC1 compares favorably with existing axis-parallel algorithms.

6. The first run through the algorithm at each node always begins at the location of the best axis-parallel hyperplane; all subsequent restarts begin at random locations.
7. Sometimes a simple axis-parallel split is preferable to an oblique split, even if the oblique split has slightly lower impurity. The user can specify such a bias as an input parameter to OC1.

## 3.3 Other Details

### 3.3.1 Impurity Measures

OC1 attempts to divide the $d$-dimensional attribute space into homogeneous regions; i.e., regions that contain examples from just one category. The goal of adding new nodes to a tree is to split up the sample space so as to minimize the "impurity" of the training set. Some algorithms measure "goodness" instead of impurity, the difference being that goodness values should be maximized while impurity should be minimized. Many different measures of impurity have been studied (Breiman et al., 1984; Quinlan, 1986; Mingers, 1989b; Buntine & Niblett, 1992; Fayyad & Irani, 1992; Heath et al., 1993b).

The OC1 system is designed to work with a large class of impurity measures. Stated simply, if the impurity measure uses only the counts of examples belonging to every category on both sides of a split, then OC1 can use it. (See Murthy and Salzberg (1994) for ways of mapping other kinds of impurity measures to this class of impurity measures.) The user can plug in any impurity measure that fits this description. The OC1 implementation includes six impurity measures, namely:

1. Information Gain
2. The Gini Index
3. The Twoing Rule
4. Max Minority
5. Sum Minority
6. Sum of Variances

Though all six of the measures have been defined elsewhere in the literature, in some cases we have made slight modifications that are defined precisely in Appendix B. Our experiments indicated that, on average, Information Gain, Gini Index and the Twoing Rule perform better than the other three measures for both axis-parallel and oblique trees. The Twoing Rule is the current default impurity measure for OC1, and it was used in all of the experiments reported in Section 4. There are, however, artificial data sets for which Sum Minority and/or Max Minority perform much better than the rest of the measures. For instance, Sum Minority easily induces the exact tree for the POL data set described in Section 4.3.1, while all other methods have difficulty finding the best tree.

**Twoing Rule.** The Twoing Rule was first proposed by Breiman et al. (1984). The value to be computed is defined as:

$$\text{TwoingValue} = (|T_L|/n) * (|T_R|/n) * (\sum_{i=1}^{k} |L_i/|T_L| - R_i/|T_R||)^2$$

where $|T_L|$ ($|T_R|$) is the number of examples on the left (right) of a split at node $T$, $n$ is the number of examples at node $T$, and $L_i$ ($R_i$) is the number of examples in category $i$ on the left (right) of the split. The TwoingValue is actually a goodness measure rather than an impurity measure. Therefore OC1 attempts to minimize the reciprocal of this value.

The remaining five impurity measures implemented in OC1 are defined in Appendix B.

#### 3.3.2 Pruning

Virtually all decision tree induction systems prune the trees they create in order to avoid overfitting the data. Many studies have found that judicious pruning results in both smaller and more accurate classifiers, for decision trees as well as other types of machine learning systems (Quinlan, 1987; Niblett, 1986; Cestnik, Kononenko, & Bratko, 1987; Kodratoff & Manago, 1987; Cohen, 1993; Hassibi & Stork, 1993; Wolpert, 1992; Schaffer, 1993). For the OC1 system we implemented an existing pruning method, but note that any tree pruning method will work fine within OC1. Based on the experimental evaluations of Mingers (1989a) and other work cited above, we chose Breiman et al.'s Cost Complexity (CC) pruning (1984) as the default pruning method for OC1. This method, which is also called Error Complexity or Weakest Link pruning, requires a separate pruning set. The pruning set can be a randomly chosen subset of the training set, or it can be approximated using cross validation. OC1 randomly chooses 10% (the default value) of the training data to use for pruning. In the experiments reported below, we only used this default value.

Briefly, the idea behind CC pruning is to create a set of trees of decreasing size from the original, complete tree. All these trees are used to classify the pruning set, and accuracy is estimated from that. CC pruning then chooses the smallest tree whose accuracy is within $k$ standard errors squared of the best accuracy obtained. When the 0-SE rule ($k = 0$) is used, the tree with highest accuracy on the pruning set is selected. When $k > 0$, smaller tree size is preferred over higher accuracy. For details of Cost Complexity pruning, see Breiman et al. (1984) or Mingers (1989a).

#### 3.3.3 Irrelevant attributes

Irrelevant attributes pose a significant problem for most machine learning methods (Breiman et al., 1984; Aha, 1990; Almuallin & Dietterich, 1991; Kira & Rendell, 1992; Salzberg, 1992; Cardie, 1993; Schlimmer, 1993; Langley & Sage, 1993; Brodley & Utgoff, 1994). Decision tree algorithms, even axis-parallel ones, can be confused by too many irrelevant attributes. Because oblique decision trees learn the coefficients of each attribute at a DT node, one might hope that the values chosen for each coefficient would reflect the relative importance of the corresponding attributes. Clearly, though, the process of searching for good coefficient values will be much more efficient when there are fewer attributes; the search space is much smaller. For this reason, oblique DT induction methods can benefit substantially by using a feature selection method (an algorithm that selects a subset of the original attribute set) in conjunction with the coefficient learning algorithm (Breiman et al., 1984; Brodley & Utgoff, 1994).

Currently, OC1 does not have a built-in mechanism to select relevant attributes. However, it is easy to include any of several standard methods (e.g., stepwise forward selection or stepwise backward selection) or even an ad hoc method to select features before running the tree-building process. For example, in separate experiments on data from the Hubble Space Telescope (Salzberg, Chandar, Ford, Murthy, & White, 1994), we used feature selection methods as a preprocessing step to OC1, and reduced the number of attributes from 20 to 2. The resulting decision trees were both simpler and more accurate. Work is currently underway to incorporate an efficient feature selection technique into the OC1 system.

Regarding missing values, if an example is missing a value for any attribute, OC1 uses the mean value for that attribute. One can of course use other techniques for handling missing values, but those were not considered in this study.

## 4. Experiments

In this section, we present two sets of experiments to support the following two claims.

1. OC1 compares favorably over a variety of real-world domains with several existing axis-parallel and oblique decision tree induction methods.
2. Randomization, both in the form of multiple local searches and random jumps, improves the quality of decision trees produced by OC1.

The experimental method used for all the experiments is described in Section 4.1. Sections 4.2 and 4.3 describe experiments corresponding to the above two claims. Each experimental section begins with a description of the data sets, and then presents the experimental results and discussion.

### 4.1 Experimental Method

We used five-fold cross validation (CV) in all our experiments to estimate classification accuracy. A $k$-fold CV experiment consists of the following steps.

1. Randomly divide the data into $k$ equal-sized disjoint partitions.
2. For each partition, build a decision tree using all data outside the partition, and test the tree on the data in the partition.
3. Sum the number of correct classifications of the $k$ trees and divide by the total number of instances to compute the classification accuracy. Report this accuracy and the average size of the $k$ trees.

Each entry in Tables 1 and 2 is a result of ten 5-fold CV experiments; i.e., the result of tests that used 50 decision trees. Each of the ten 5-fold cross validations used a different random partitioning of the data. Each entry in the tables reports the mean and standard deviation of the classification accuracy, followed by the mean and standard deviation of the decision tree size (measured as the number of leaf nodes). Good results should have high values for accuracy, low values for tree size, and small standard deviations.

In addition to OC1, we also included in the experiments an axis-parallel version of OC1, which only considers axis-parallel hyperplanes. We call this version, described in Section 3.2, OC1-AP. In all our experiments, both OC1 and OC1-AP used the Twoing Rule (Section 3.3.1) to measure impurity. Other parameters to OC1 took their default values unless stated otherwise. (Defaults include the following: number of restarts at each node: 20. Number of random jumps attempted at each local minimum: 5. Order of coefficient perturbation: Sequential. Pruning method: Cost Complexity with the 0-SE rule, using 10% of the training set exclusively for pruning.)

In our comparison, we used the oblique version of the CART algorithm, CART-LC. We implemented our own version of CART-LC, following the description in Breiman et al. (1984, Chapter 5); however, there may be differences between our version and other versions of this system (note that CART-LC is not freely available). Our implementation of CART-LC measured impurity with the Twoing Rule and used 0-SE Cost Complexity pruning with a separate test set, just as OC1 does. We did not include any feature selection methods in CART-LC or in OC1, and we did not implement normalization. Because the CART coefficient perturbation algorithm may alternate indefinitely between two locations of a hyperplane (see Section 2), we imposed an arbitrary limit of 100 such perturbations before forcing the perturbation algorithm to halt.

We also included axis-parallel CART and C4.5 in our comparisons. We used the implementations of these algorithms from the IND 2.1 package (Buntine, 1992). The default cart0 and c4.5 "styles" defined in the package were used, without altering any parameter settings. The cart0 style uses the Twoing Rule and 0-SE cost complexity pruning with 10-fold cross validation. The pruning method, impurity measure and other defaults of the c4.5 style are the same as those described in Quinlan (1993a).

### 4.2 OC1 vs. Other Decision Tree Induction Methods

Table 1 compares the performance of OC1 to three well-known decision tree induction methods plus OC1-AP on six different real-world data sets. In the next section we will consider artificial data, for which the concept definition can be precisely characterized.

#### 4.2.1 Description of Data Sets

**Star/Galaxy Discrimination.** Two of our data sets came from a large set of astronomical images collected by Odewahn et al. (Odewahn, Stockwell, Pennington, Humphreys, & Zumach, 1992). In their study, they used these images to train artificial neural networks running the perceptron and back propagation algorithms. The goal was to classify each example as either "star" or "galaxy." Each image is characterized by 14 real-valued attributes, where the attributes were measurements defined by astronomers as likely to be relevant for this task. The objects in the image were divided by Odewahn et al. into "bright" and "dim" data sets based on the image intensity values, where the dim images are inherently more difficult to classify. (Note that the "bright" objects are only bright in relation to others in this data set. In actuality they are extremely faint, visible only to the most powerful telescopes.) The bright set contains 2462 objects and the dim set contains 4192 objects.

In addition to the results reported in Table 1, the following results have appeared on the Star/Galaxy data. Odewahn et al. (1992) reported accuracy of 99.8% accuracy on the bright objects, and 92.0% on the dim ones, although it should be noted that this study used a single training and test set partition. Heath (1992) reported 99.0% accuracy on the bright objects using SADT, with an average tree size of 7.03 leaves. This study also used a single training and test set. Salzberg (1992) reported accuracies of 98.8% on the bright objects, and 95.1% on the dim objects, using 1-Nearest Neighbor (1-NN) coupled with a feature selection method that reduces the number of features.

**Breast Cancer Diagnosis.** Mangasarian and Bennett have compiled data on the problem of diagnosing breast cancer to test several new classification methods (Mangasarian et al., 1990; Bennett & Mangasarian, 1992, 1994a). This data represents a set of patients with breast cancer, where each patient was characterized by nine numeric attributes plus the diagnosis of the tumor as benign or malignant. The data set currently has 683 entries and is available from the UC Irvine machine learning repository (Murphy & Aha, 1994). Heath et al. (1993b) reported 94.9% accuracy on a subset of this data set (it then had only 470 instances), with an average decision tree size of 4.6 nodes, using SADT. Salzberg (1991) reported 96.0% accuracy using 1-NN on the same (smaller) data set. Herman and Yeung (1992) reported 99.0% accuracy using piece-wise linear classification, again using a somewhat smaller data set.

| *Algorithm* | Bright S/G | Dim S/G | Cancer | Iris | Housing | Diabetes |
|---|---|---|---|---|---|---|
| OC1 | **98.9**±0.2 | **95.0**±0.3 | **96.2**±0.3 | 94.7±3.1 | 82.4±0.8 | **74.4**±1.0 |
| | 4.3±1.0 | 13.0±8.7 | 2.8±0.9 | 3.1±0.2 | 6.9±3.2 | 5.4±3.8 |
| CART-LC | 98.8±0.2 | 92.8±0.5 | 95.3±0.6 | 93.5±2.9 | 81.4±1.2 | 73.7±1.2 |
| | 3.9±1.3 | 24.2±8.7 | 3.5±0.9 | 3.2±0.3 | 5.8±3.2 | 8.0±5.2 |
| OC1-AP | 98.1±0.2 | 94.0±0.2 | 94.5±0.5 | 92.7±2.4 | 81.8±1.0 | 73.8±1.0 |
| | 6.9±2.4 | 29.3±8.8 | 6.4±1.7 | 3.2±0.3 | 8.6±4.5 | 11.4±7.5 |
| CART-AP | 98.5±0.5 | 94.2±0.7 | 95.0±1.6 | 93.8±3.7 | 82.1±3.5 | 73.9±3.4 |
| | 13.9±5.7 | 30.4±10 | 11.5±7.2 | 4.3±1.6 | 15.1±10 | 11.5±9.1 |
| C4.5 | 98.5±0.5 | 93.3±0.8 | 95.3±2.0 | **95.1**±3.2 | **83.2**±3.1 | 71.4±3.3 |
| | 14.3±2.2 | 77.9±7.4 | 9.8±2.2 | 4.6±0.8 | 28.2±3.3 | 56.3±7.9 |

Table 1: Comparison of OC1 and other decision tree induction methods on six different data sets. The first line for each method gives accuracies, and the second line gives average tree sizes. The highest accuracy for each domain appears in boldface.

**Classifying Irises.** This is Fisher's famous iris data, which has been extensively studied in the statistics and machine learning literature. The data consists of 150 examples, where each example is described by four numeric attributes. There are 50 examples of each of three different types of iris flower. Weiss and Kapouleas (1989) obtained accuracies of 96.7% and 96.0% on this data with back propagation and 1-NN, respectively.

**Housing Costs in Boston.** This data set, also available as a part of the UCI ML repository, describes housing values in the suburbs of Boston as a function of 12 continuous attributes and 1 binary attribute (Harrison & Rubinfeld, 1978). The category variable (median value of owner-occupied homes) is actually continuous, but we discretized it so that category = 1 if value $<$ \$21000, and 2 otherwise. For other uses of this data, see (Belsley, 1980; Quinlan, 1993b).

**Diabetes diagnosis.** This data catalogs the presence or absence of diabetes among Pima Indian females, 21 years or older, as a function of eight numeric-valued attributes. The original source of the data is the National Institute of Diabetes and Digestive and Kidney Diseases, and it is now available in the UCI repository. Smith et al. (1988) reported 76% accuracy on this data using their ADAP learning algorithm, using a different experimental method from that used here.

#### 4.2.2 Discussion

The table shows that, for the six data sets considered here, OC1 consistently finds better trees than the original oblique CART method. Its accuracy was greater in all six domains, although the difference was significant (more than 2 standard deviations) only for the dim star/galaxy problem. The average tree sizes were roughly equal for five of the six domains, and for the dim stars and galaxies, OC1 found considerably smaller trees. These differences will be analyzed and quantified further by using artificial data, in the following section.

Out of the five decision tree induction methods, OC1 has the highest accuracy on four of the six domains: bright stars, dim stars, cancer diagnosis, and diabetes diagnosis. On the remaining two domains, OC1 has the second highest accuracy in each case. Not surprisingly, the oblique methods (OC1 and CART-LC) generally find much smaller trees than the axis-parallel methods. This difference can be quite striking for some domains—note, for example, that OC1 produced a tree with just 13 nodes on average for the dim star/galaxy problem, while C4.5 produced a tree with 78 nodes, 6 times larger. Of course, in domains for which an axis-parallel tree is the appropriate representation, axis-parallel methods should compare well with oblique methods in terms of tree size. In fact, for the Iris data, all the methods found similar-sized trees.

### 4.3 Randomization Helps OC1

In our second set of experiments, we examine more closely the effect of introducing randomized steps into the algorithm for finding oblique splits. Our experiments demonstrate that OC1's ability to produce an accurate tree from a set of training data is clearly enhanced by the two kinds of randomization it uses. More precisely, we use three artificial data sets (for which the underlying concept is known to the experimenters) to show that OC1's performance improves substantially when the deterministic hill climbing is augmented in any of three ways:

- with multiple restarts from random initial locations,
- with perturbations in random directions at local minima, or
- with both of the above randomization steps.

In order to find clear differences between algorithms, one needs to know that the concept underlying the data is indeed difficult to learn. For simple concepts (say, two linearly separable classes in 2-D), many different learning algorithms will produce very accurate classifiers, and therefore the advantages of randomization may not be detectable. It is known that many of the commonly-used data sets from the UCI repository are easy to learn with very simple representations (Holte, 1993); therefore those data sets may not be ideal for our purposes. Thus we created a number of artificial data sets that present different problems for learning, and for which we know the "correct" concept definition. This allows us to quantify more precisely how the parameters of our algorithm affect its performance.

A second purpose of this experiment is to compare OC1's search strategy with that of two existing oblique decision tree induction systems – LMDT (Brodley & Utgoff, 1992) and SADT (Heath et al., 1993b). We show that the quality of trees induced by OC1 is as good as, if not better than, that of the trees induced by these existing systems on three artificial domains. We also show that OC1 achieves a good balance between amount of effort expended in search and the quality of the tree induced.

Both LMDT and SADT used information gain for this experiment. However, we did not change OC1's default measure (the Twoing Rule) because we observed, in experiments not reported here, that OC1 with information gain does not produce significantly different results. The maximum number of successive, unproductive perturbations allowed at any node was set at 10000 for SADT. For all other parameters, we used default settings provided with the systems.

#### 4.3.1 Description of Artificial Data

**LS10** The LS10 data set has 2000 instances divided into two categories. Each instance is described by ten attributes $x_1,\ldots,x_{10}$, whose values are uniformly distributed in the range [0,1]. The data is linearly separable with a 10-D hyperplane (thus the name LS10) defined by the equation $x_1 + x_2 + x_3 + x_4 + x_5 < x_6 + x_7 + x_8 + x_9 + x_{10}$. The instances were all generated randomly and labelled according to which side of this hyperplane they fell on. Because oblique DT induction methods intuitively should prefer a linear separator if one exists, it is interesting to compare the various search techniques on this data set where we know a separator exists. The task is relatively simple for lower dimensions, so we chose 10-dimensional data to make it more difficult.

**POL** This data set is shown in Figure 9. It has 2000 instances in two dimensions, again divided into two categories. The underlying concept is a set of four parallel oblique lines (thus the name POL), dividing the instances into five homogeneous regions. This concept is more difficult to learn than a single linear separator, but the minimal-size tree is still quite small.

**RCB** RCB stands for "rotated checker board"; this data set has been the subject of other experiments on hard classification problems for decision trees (Murthy & Salzberg, 1994). The data set, shown in Figure 9, has 2000 instances in 2-D, each belonging to one of eight categories. This concept is difficult to learn for any axis-parallel method, for obvious reasons. It is also quite difficult for oblique methods, for several reasons. The biggest problem is that the "correct" root node, as shown in the figure, does not separate out any class by itself. Some impurity measures (such as Sum Minority) will fail miserably on this problem, although others (e.g., the Twoing Rule) work much better. Another problem is that a deterministic coefficient perturbation algorithm can get stuck in local minima in many places on this data set.

Table 2 summarizes the results of this experiment in three smaller tables, one for each data set. In each smaller table, we compare four variants of OC1 with LMDT and SADT. The different results for OC1 were obtained by varying both the number of restarts and the number of random jumps. When random jumps were used, up to twenty random jumps were tried at each local minimum. As soon as one was found that improved the impurity of the current hyperplane, the algorithm moved the hyperplane and started running the deterministic perturbation procedure again. If none of the 20 random jumps improved the impurity, the search halted and further restarts (if any) were tried. The same training and test partitions were used for all methods for each cross-validation run (recall that the results are an average of ten 5-fold CVs). The trees were not pruned for any of the algorithms, because the data were noise-free and furthermore the emphasis was on search.

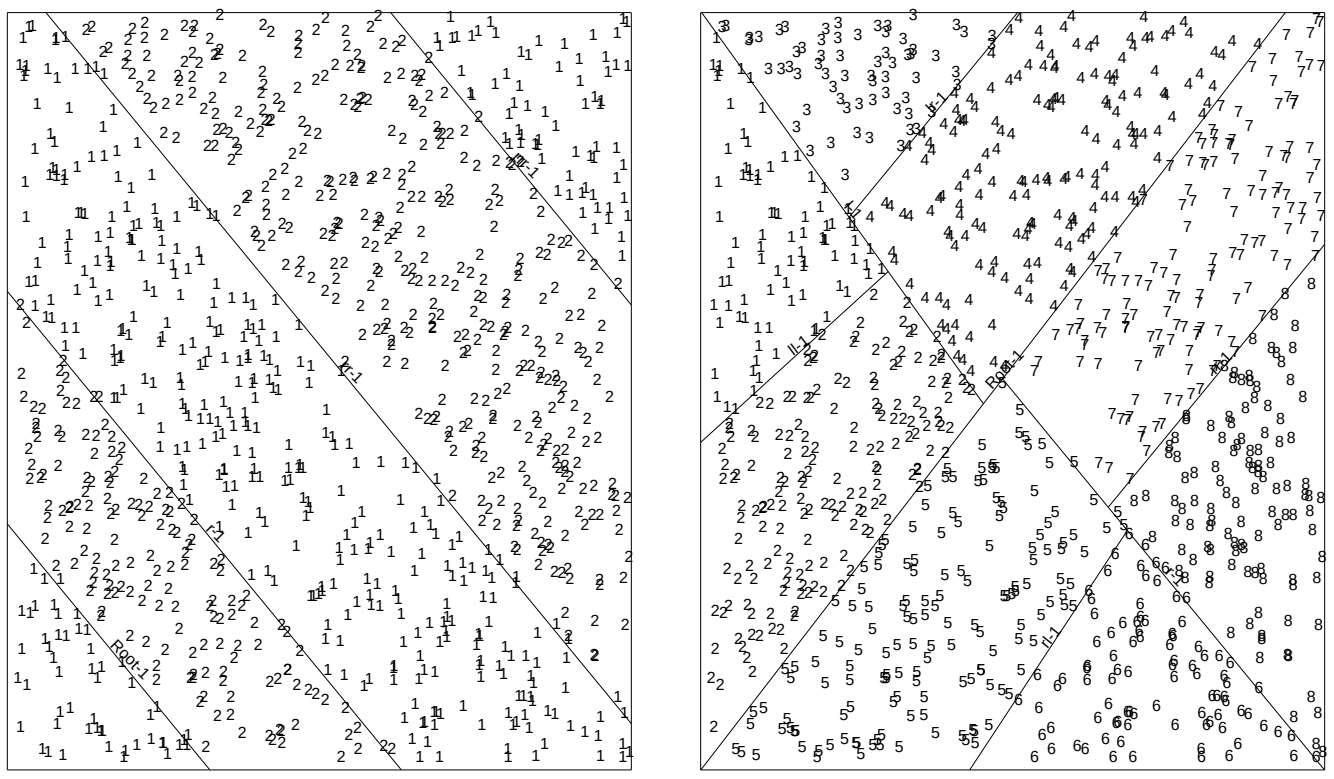

Figure 9: The POL and RCB data sets

| Linearly Separable 10-D (LS10) data | | | |
|---|---|---|---|
| R:J | Accuracy | Size | Hyperplanes |
| 0:0 | 89.8±1.2 | 67.0±5.8 | 2756 |
| 0:20 | 91.5±1.5 | 55.2±7.0 | 3824 |
| 20:0 | 95.0±0.6 | 25.6±2.4 | 24913 |
| 20:20 | 97.2±0.7 | 13.9±3.2 | 30366 |
| LMDT | 99.7±0.2 | 2.2±0.5 | 9089 |
| SADT | 95.2±1.8 | 15.5±5.7 | 349067 |
| **Parallel Oblique Lines (POL) data** | | | |
| R:J | Accuracy | Size | Hyperplanes |
| 0:0 | 98.3±0.3 | 21.6±1.9 | 164 |
| 0:20 | 99.3±0.2 | 9.0±1.0 | 360 |
| 20:0 | 99.1±0.2 | 14.2±1.1 | 3230 |
| 20:20 | 99.6±0.1 | 5.5±0.3 | 4852 |
| LMDT | 89.6±10.2 | 41.9±19.2 | 1732 |
| SADT | 99.3±0.4 | 8.4±2.1 | 85594 |
| **Rotated Checker Board (RCB) data** | | | |
| R:J | Accuracy | Size | Hyperplanes |
| 0:0 | 98.4±0.2 | 35.5±1.4 | 573 |
| 0:20 | 99.3±0.3 | 19.7±0.8 | 1778 |
| 20:0 | 99.6±0.2 | 12.0±1.4 | 6436 |
| 20:20 | 99.8±0.1 | 8.7±0.4 | 11634 |
| LMDT | 95.7±2.3 | 70.1±9.6 | 2451 |
| SADT | 97.9±1.1 | 32.5±4.9 | 359112 |

Table 2: The effect of randomization in OC1. The first column, labelled R:J, shows the number of restarts (R) followed by the maximum number of random jumps (J) attempted by OC1 at each local minimum. Results with LMDT and SADT are included for comparison after the four variants of OC1. Size is average tree size measured by the number of leaf nodes. The third column shows the average number of hyperplanes each algorithm considered while building one tree.

Table 2 also includes the number of hyperplanes considered by each algorithm while building a complete tree. Note that for OC1 and SADT, the number of hyperplanes considered is generally much larger than the number of perturbations actually made, because both these algorithms compare newly generated hyperplanes to existing hyperplanes before adjusting an existing one. Nevertheless, this number is a good estimate of much effort each algorithm expends, because every new hyperplane must be evaluated according to the impurity measure. For LMDT, the number of hyperplanes considered is identical to the actual number of perturbations.

#### 4.3.2 Discussion

The OC1 results here are quite clear. The first line of each table, labelled 0:0, gives the accuracies and tree sizes when no randomization is used — this variant is very similar to the CART-LC algorithm. As we increase the use of randomization, accuracy increases while tree size decreases, which is exactly the result we had hoped for when we decided to introduce randomization into the method.

Looking more closely at the tables, we can ask about the effect of random jumps alone. This is illustrated in the second line (0:20) of each table, which attempted up to 20 random jumps at each local minimum and no restarts. Accuracy increased by 1-2% on each domain, and tree size decreased dramatically, roughly by a factor of two, in the POL and RCB domains. Note that because there is no noise in these domains, very high accuracies should be expected. Thus increases of more than a few percent in accuracy are not possible.

Looking at the third line of each sub-table in Table 2, we see the effect of multiple restarts on OC1. With 20 restarts but no random jumps to escape local minima, the improvement is even more noticeable for the LS10 data than when random jumps alone were used. For this data set, accuracy jumped significantly, from 89.8 to 95.0%, while tree size dropped from 67 to 26 nodes. For the POL and RCB data, the improvements were comparable to those obtained with random jumps. For the RCB data, tree size dropped by a factor of 3 (from 36 leaf nodes to 12 leaf nodes) while accuracy increased from 98.4 to 99.6%.

The fourth line of each table shows the effect of both the randomized steps. Among the OC1 entries, this line has both the highest accuracies and the smallest trees for all three data sets, so it is clear that randomization is a big win for these kinds of problems. In addition, note that the smallest tree for the RCB data should have eight leaf nodes, and OC1's average trees, without pruning, had just 8.7 leaf nodes. It is clear that for this data set, which we thought was the most difficult one, OC1 came very close to finding the optimal tree on nearly every run. (Recall that numbers in the table are the average of 10 5-fold CV experiments; i.e., an average of 50 decision trees.) The LS10 data show how difficult it can be to find a very simple concept in higher dimensions—the optimal tree there is just a single hyperplane (two nodes), but OC1 was unable to find it with the current parameter settings.[8] The POL data required a minimum of 5 leaf nodes, and OC1 found this minimal-size tree most of the time, as can be seen from the table. Although not shown in the Table,

8. In a separate experiment, we found that OC1 consistently finds the linear separator for the LS10 data when 10 restarts and 200 random jumps are used.

OC1 using Sum Minority performed better for the POL data than the Twoing Rule or any other impurity measure; i.e., it found the correct tree using less time.

The results of LMDT and SADT on this data lead to some interesting insights. Not surprisingly, LMDT does very well on the linearly separable (LS10) data, and does not require an inordinate amount of search. Clearly, if the data is linearly separable, one should use a method such as LMDT or linear programming. OC1 and SADT have difficulty finding the linear separator, although in our experiments OC1 did eventually find it, given sufficient time.

On the other hand, for both of the non-linearly separable data sets, LMDT produces much larger trees that are significantly less accurate than those produced by OC1 and SADT. Even the deterministic variant of OC1 (using zero restarts and zero random jumps) outperforms LMDT on these problems, with much less search.

Although SADT sometimes produces very accurate trees, its main weakness was the enormous amount of search time it required, roughly 10-20 times greater than OC1 even using the 20:20 setting. One explanation of OC1's advantage is its use of directed search, as opposed to the strictly random search used by simulated annealing. Overall, Table 2 shows that OC1's use of randomization was quite effective for the non-linearly separable data.

It is natural to ask *why* randomization helps OC1 in the task of inducing decision trees. Researchers in combinatorial optimization have observed that randomized search usually succeeds when the search space holds an abundance of good solutions (Gupta, Smolka, & Bhaskar, 1994). Furthermore, randomization can improve upon deterministic search when many of the local maxima in a search space lead to poor solutions. In OC1's search space, a local maximum is a hyperplane that cannot be improved by the deterministic search procedure, and a "solution" is a complete decision tree. If a significant fraction of local maxima lead to bad trees, then algorithms that stop at the first local maximum they encounter will perform poorly. Because randomization allows OC1 to consider many different local maxima, if a modest percentage of these maxima lead to good trees, then it has a good chance of finding one of those trees. Our experiments with OC1 thus far indicate that the space of oblique hyperplanes usually contains numerous local maxima, and that a substantial percentage of these locally good hyperplanes lead to good decision trees.

## 5. Conclusions and Future Work

This paper has described OC1, a new system for constructing oblique decision trees. We have shown experimentally that OC1 can produce good classifiers for a range of real-world and artificial domains. We have also shown how the use of randomization improves upon the original algorithm proposed by Breiman et al. (1984), without significantly increasing the computational cost of the algorithm.

The use of randomization might also be beneficial for axis-parallel tree methods. Note that although they do find the optimal test (with respect to an impurity measure) for each node of a tree, the complete tree may not be optimal: as is well known, the problem of finding the smallest tree is NP-Complete (Hyafil & Rivest, 1976). Thus even axis-parallel decision tree methods do not produce "ideal" decision trees. Quinlan has suggested that his windowing algorithm might be used as a way of introducing randomization into C4.5, even though the algorithm was designed for another purpose (Quinlan, 1993a). (The windowing algorithm selects a random subset of the training data and builds a tree using that.) We believe that randomization is a powerful tool in the context of decision trees, and our experiments are just one example of how it might be exploited. We are in the process of conducting further experiments to quantify more accurately the effects of different forms of randomization.

It should be clear that the ability to produce oblique splits at a node broadens the capabilities of decision tree algorithms, especially as regards domains with numeric attributes. Of course, axis-parallel splits are simpler, in the sense that the description of the split only uses one attribute at each node. OC1 uses oblique splits only when their impurity is less than the impurity of the best axis-parallel split; however, one could easily penalize the additional complexity of an oblique split further. This remains an open area for further research. A more general point is that if the domain is best captured by a tree that uses oblique hyperplanes, it is desirable to have a system that can generate that tree. We have shown that for some problems, including those used in our experiments, OC1 builds small decision trees that capture the domain well.

## Appendix A. Complexity Analysis of OC1

In the following, we show that OC1 runs efficiently even in the worst case. For a data set with $n$ examples (points) and $d$ attributes per example, OC1 uses at most $O(dn^2 \log n)$ time. We assume $n > d$ for our analysis.

For the analysis here, we assume the coefficients of a hyperplane are adjusted in sequential order (the Seq method described in the paper). The number of restarts at a node will be $r$, and the number of random jumps tried will be $j$. Both $r$ and $j$ are constants, fixed in advance of running the algorithm.

Initializing the hyperplane to a random position takes just $O(d)$ time. We need to consider first the maximum amount of work OC1 can do before it finds a new location for the hyperplane. Then we need to consider how many times it can move the hyperplane.

1. Attempting to perturb the first coefficient ($a_1$) takes $O(dn + n \log n)$ time. Computing $U_i$'s for all the points (equation 2) requires $O(dn)$ time, and sorting the $U_i$'s takes $O(n \log n)$. This gives us $O(dn + n \log n)$ work.

2. If perturbing $a_1$ does not improve things, we try to perturb $a_2$. Computing all the new $U_i$'s will take just $O(n)$ time because only one term is different for each $U_i$. Re-sorting will take $O(n \log n)$, so this step takes $O(n) + O(n \log n) = O(n \log n)$ time.

3. Likewise $a_3, \ldots, a_d$ will each take $O(n \log n)$ additional time, assuming we still have not found a better hyperplane after checking each coefficient. Thus the total time to cycle through and attempt to perturb all these additional coefficients is $(d-1) * O(n \log n) = O(dn \log n)$.

4. Summing up, the time to cycle through all coefficients is $O(dn \log n) + O(dn + n \log n) = O(dn \log n)$.

5. If none of the coefficients improved the split, then we attempt to make up to $j$ random jumps. Since $j$ is a constant, we will just consider $j = 1$ for our analysis. This step involves choosing a random vector and running the perturbation algorithm to solve for $\alpha$, as explained in Section 3.2. As before, we need to compute a set of $U_i$'s and sort them, which takes $O(dn + n \log n)$ time. Because this amount of time is dominated by the time to adjust all the coefficients, the total time so far is still $O(dn \log n)$. This is the most time OC1 can spend at a node before either halting or finding an improved hyperplane.

6. Assuming OC1 is using the Sum Minority or Max Minority error measure, it can only reduce the impurity of the hyperplane $n$ times. This is clear because each improvement means one more example will be correctly classified by the new hyperplane. Thus the total amount of work at a node is limited to $n * O(dn \log n) = O(dn^2 \log n)$. (This analysis extends, with at most linear cost factors, to Information Gain, Gini Index and Twoing Rule when there are two categories. It will not apply to a measure that, for example, uses the distances of mis-classified objects to the hyperplane.) In practice, we have found that the number of improvements per node is much smaller than $n$.

Assuming that OC1 only adjusts a hyperplane when it improves the impurity measure, it will do $O(dn^2 \log n)$ work in the worst case.

However, OC1 allows a certain number of adjustments to the hyperplane that do not improve the impurity, although it will never accept a change that worsens the impurity. The number allowed is determined by a constant known as "stagnant-perturbations". Let this value be $s$. This works as follows.

Each time OC1 finds a new hyperplane that improves on the old one, it resets a counter to zero. It will move the new hyperplane to a different location that has *equal* impurity at most $s$ times. After each of these moves it repeats the perturbation algorithm. Whenever impurity is reduced, it re-starts the counter and again allows $s$ moves to equally good locations. Thus it is clear that this feature just increases the worst-case complexity of OC1 by a constant factor, $s$.

Finally, note that the overall cost of OC1 is also $O(dn^2 \log n)$, i.e., this is an upper bound on the total running time of OC1 independent of the size of the tree it ends up creating. (This upper bound applies to Sum Minority and Max Minority; an open question is whether a similar upper bound can be proven for Information Gain or the Gini Index.) Thus the worst-case asymptotic complexity of our system is comparable to that of systems that construct axis-parallel decision trees, which have $O(dn^2)$ worst-case complexity. To sketch the intuition that leads to this bound, let $G$ be the total impurity summed over all leaves in a partially constructed tree (i.e., the sum of currently misclassified points in the tree). Now observe that each time we run the perturbation algorithm on any node in the tree, we either halt or improve $G$ by at least one unit. The worst-case analysis for one node is realized when the perturbation algorithm is run once for every one of the $n$ examples, but when this happens, there would no longer be any mis-classified examples and the tree would be complete.

## Appendix B. Definitions of impurity measures available in OC1

In addition to the Twoing Rule defined in the text, OC1 contains built-in definitions of five additional impurity measures, defined as follows. In each of the following definitions, the set of examples $T$ at the node about to be split contains $n$ ($> 0$) instances that belong to one of $k$ categories. (Initially this set is the entire training set.) A hyperplane $H$ divides $T$ into two non-overlapping subsets $T_L$ and $T_R$ (i.e., left and right). $L_j$ and $R_j$ are the number of instances of category $j$ in $T_L$ and $T_R$ respectively. All the impurity measures initially check to see if $T_L$ and $T_R$ are homogeneous (i.e., all examples belong to the same category), and if so return minimum (zero) impurity.

**Information Gain.** This measure of information gained from a particular split was popularized in the context of decision trees by Quinlan (1986). Quinlan's definition makes information gain a goodness measure; i.e., something to maximize. Because OC1 attempts to minimize whatever impurity measure it uses, we use the reciprocal of the standard value of information gain in the OC1 implementation.

**Gini Index.** The Gini Criterion (or Index) was proposed for decision trees by Breiman et al. (1984). The Gini Index as originally defined measures the probability of misclassification of a set of instances, rather than the impurity of a split. We implement the following variation:

$$\text{GiniL} = 1.0 - \sum_{i=1}^{k} (L_i/|T_L|)^2$$

$$\text{GiniR} = 1.0 - \sum_{i=1}^{k} (R_i/|T_R|)^2$$

$$\text{Impurity} = (|T_L| * \text{GiniL} + |T_R| * \text{GiniR})/n$$

where GiniL is the Gini Index on the "left" side of the hyperplane and GiniR is that on the right.

**Max Minority.** The measures Max Minority, Sum Minority and Sum Of Variances were defined in the context of decision trees by Heath, Kasif, and Salzberg (1993b).[9] Max Minority has the theoretical advantage that a tree built minimizing this measure will have depth at most $\log n$. Our experiments indicated that this is not a great advantage in practice: seldom do other impurity measures produce trees substantially deeper than those produced with Max Minority. The definition is:

$$\text{MinorityL} = \sum_{i=1, i \neq \max L_i}^{k} L_i$$

$$\text{MinorityR} = \sum_{i=1, i \neq \max R_i}^{k} R_i$$

$$\text{Max Minority} = \max(\text{MinorityL}, \text{MinorityR})$$

9. Sum Of Variances was called Sum of Impurities by Heath et al.

**Sum Minority.** This measure is very similar to Max Minority. If MinorityL and MinorityR are defined as for the Max Minority measure, then Sum Minority is just the sum of these two values. This measure is the simplest way of quantifying impurity, as it simply counts the number of misclassified instances.

Though Sum Minority performs well on some domains, it has some obvious flaws. As one example, consider a domain in which $n = 100, d = 1$, and $k = 2$ (i.e., 100 examples, 1 numeric attribute, 2 classes). Suppose that when the examples are sorted according to the single attribute, the first 50 instances belong to category 1, followed by 24 instances of category 2, followed by 26 instances of category 1. Then *all* possible splits for this distribution have a sum minority of 24. Therefore it is impossible when using Sum Minority to distinguish which split is preferable, although splitting at the alternations between categories is clearly better.

**Sum Of Variances.** The definition of this measure is:

$$\text{VarianceL} = \sum_{i=1}^{|T_L|} (Cat(T_{L_i}) - \sum_{j=1}^{|T_L|} Cat(T_{L_j})/|T_L|)^2$$

$$\text{VarianceR} = \sum_{i=1}^{|T_R|} (Cat(T_{R_i}) - \sum_{j=1}^{|T_R|} Cat(T_{R_j})/|T_R|)^2$$

$$\text{Sum of Variances} = \text{VarianceL} + \text{VarianceR}$$

where Cat($T_i$) is the category of instance $T_i$. As this measure is computed using the actual class labels, it is easy to see that the impurity computed varies depending on how numbers are assigned to the classes. For instance, if $T_1$ consists of 10 points of category 1 and 3 points of category 2, and if $T_2$ consists of 10 points of category 1 and 3 points of category 5, then the Sum Of Variances values are different for $T_1$ and $T_2$. To avoid this problem, OC1 uniformly reassigns category numbers according to the frequency of occurrence of each category at a node before computing the Sum Of Variances.

## Acknowledgements

The authors thank Richard Beigel of Yale University for suggesting the idea of jumping in a random direction. Thanks to Wray Buntine of Nasa Ames Research Center for providing the IND 2.1 package, to Carla Brodley for providing the LMDT code, and to David Heath for providing the SADT code and for assisting us in using it. Thanks also to three anonymous reviewers for many helpful suggestions. This material is based upon work supported by the National Science foundation under Grant Nos. IRI-9116843, IRI-9223591, and IRI-9220960.